\documentclass[letterpaper]{article} 
\usepackage{aaai2026}  
\usepackage{times}  
\usepackage{helvet}  
\usepackage{courier}  
\usepackage[hyphens]{url}  
\usepackage{graphicx} 
\usepackage{natbib}  
\usepackage{caption} 
\usepackage{algorithm}
\usepackage{algorithmic}
\usepackage{amsmath}
\usepackage{subfigure}
\usepackage{booktabs}
\usepackage{newfloat}
\usepackage{listings}
\DeclareCaptionStyle{ruled}{labelfont=normalfont,labelsep=colon,strut=off} 
\floatstyle{ruled}
\newfloat{listing}{tb}{lst}{}
\floatname{listing}{Listing}
\title{FT-MoE: Sustainable-learning Mixture of Experts for Fault-Tolerant Computing}
\author {
    Wenjing Xiao\textsuperscript{\rm 1,2}, 
    Wenhao Song\textsuperscript{\rm 1,2}, 
    Miaojiang Chen\textsuperscript{\rm 1,2,*}, 
    Min Chen\textsuperscript{\rm 3,4}
}
\affiliations {
    \textsuperscript{\rm 1}Guangxi Key Laboratory of Multimedia Communications and Network Technology, Nanning, 530004, China\\
    \textsuperscript{\rm 2}School of Computer, Electronics and Information, Guangxi University, Nanning, 530004, China\\
    \textsuperscript{\rm 3}School of Computer Science and Engineering, South China University of Technology, Guangzhou 510006, China\\
    \textsuperscript{\rm 4}Pazhou Laboratory, Guangzhou 510330, China\\
    wenjingx@gxu.edu.cn, wenhaos@st.gxu.edu.cn, mjchen\_cs@gxu.edu.cn, minchen@ieee.org
}

\usepackage{bibentry}

\begin{document}

\maketitle

\begin{abstract}
Intelligent fault-tolerant (FT) computing has recently demonstrated significant advantages in predicting and diagnosing faults proactively, thereby ensuring reliable service delivery. However, due to the heterogeneity of fault knowledge, dynamic workloads, and limited data support, existing deep learning-based FT algorithms face challenges in fault detection quality and training efficiency. This is primarily because their homogenization of fault knowledge perception difficuties to fully capture diverse and complex fault patterns. To address these challenges, we propose FT-MoE, a sustainable-learning fault-tolerant computing framework based on a dual-path architecture for high-accuracy fault detection and classification. This model employs a mixture-of-experts (MoE) architecture, enabling different parameters to learn distinct fault knowledge. Additionally, we adopt a two-stage learning scheme that combines comprehensive offline training with continual online tuning, allowing the model to adaptively optimize its parameters in response to evolving real-time workloads. To facilitate realistic evaluation, we construct a new fault detection and classification dataset for edge networks, comprising 10,000 intervals with fine-grained resource features, surpassing existing datasets in both scale and granularity. Finally, we conduct extensive experiments on the FT benchmark to verify the effectiveness of FT-MoE. Results demonstrate that our model outperforms state-of-the-art methods. 
\end{abstract}
\begin{links}
\link{Code}{https://github.com/1291632523/FT-MoE}
\end{links}

\section{Introduction}
With the development of Internet of Things (IoT) technology, the number of computation tasks at network edges has increased dramatically~\cite{TULI2023103648}. When edge devices simultaneously process multiple computation tasks, the limited resource of edge network leads to high device failures and low quality of experience~\cite{TMM2025}. Traditional restoring methods of replicating or resubmitting tasks is to take action after failure has been observed, and cannot prevent expensive repair costs caused by system failures~\cite{SHAIKH2024110803}. Thus, to prevent losses of event failure and system downtime, proactive methods that predict failures before they occur have received increasing attention~\cite{AHMED2024100814}.
Recently, model-based fault tolerant (FT) computing adopts specialized algorithms for intelligent task scheduling to address event failure problems and receives increasing attention~\cite{XU2024404}~\cite{10.1145/3676641.3716006}. Among them, deep learning-based FT schemes that do not require any assumptions or specialized algorithms have become the most popular solution~\cite{Assiri2025FaultTI}~\cite{AMIN2024121956}. Under the guidance of large amounts of labeled data, deep neural networks capture the feature of abnormal signals well for fault prediction and judgment. 
However, due to limited computing and communication capacity in edge network~\cite{xiao2025graphedge}, deep learning-based schemes still face difficulties in guaranteeing fault diagnose quality of service systems~\cite{HAZRA2024102460}, facing several challenges: 
(1) \textbf{High Knowledge Heterogeneity:} 
As a multi-task learning, FT task features both fault detection and fault classification. Their sub-tasks have completely different feature learning paths, and their domain knowledge varies from fault to fault~\cite{8598484}. However, existing approaches~\cite{NA2025109740}~\cite{QIAN2025111837}~\cite{Negi2021CMODLBAE} typically use same parameters and ignore the heterogeneity of fault types and learning tasks, resulting in performance inefficiency.
(2) \textbf{Highly Dynamic Environment:} 
Since tasks in edge network are complex and variable, fault detection must adapt to the dynamic environment. This requires the FT model to be not only robust, but also flexible enough to adapt dynamic environments in real time.
(3) \textbf{Lack of Realistic FT Datasets:}
Most existing datasets used in related studies have coarse-grained resource features. These datasets do not reflect the real operating conditions of multitask scheduling and resource-constrained scenarios. As a result, models trained on them struggle to generalize to real-world deployments, which limits the research of edge fault detection algorithms.

In response to the above issues, we propose FT-MoE, a dynamic fault prediction model for fault-tolerant scheduling, which aims to improve the fault detection performance. Unlike existing DL-based approaches, we propose a sustainable learning fault-tolerant model based on a dual-path architecture to achieve efficient FT computing of multiple tasks. Specifically, to learn heterogeneous knowledge from detection data, we propose a mixture-of-experts(MoE) network with efficient gating modules as the first path, enabling dynamic tuning by adjusting the number of experts based on the activation differences~\cite{10988646}. To efficiently capture the impact of task migration on device status, we adopt efficient dynamic graph attention network as the second path. Then, we use a cross multi-head attention to model the relationship of heterogeneous knowledge and task scheduling effects extracted from the two paths.
Further, to adapt dynamic service environments, FT-MoE employs continual learning. We use a two-stage model optimization scheme, consisting of offline training and online tuning. The Offline training provides enough time to adjust and optimize parameters. The online tuning supports continual learning of FT-MoE, allowing it to dynamically add or remove parameters to adapt to dynamic environments. 
In addition, to address the lack of realistic fault tolerance datasets, we propose a dataset with fine-grained host resource features and scheduling decisions that better capture the dynamics of real-world environments. 
Overall, our main contributions are summarized as follows:
\begin{itemize}
    \item We propose FT-MoE, a dual-path fault-tolerant model that integrates task-adaptive expert selection MoE and dynamic graph attention. The first path uses a task-adaptive MoE to handle heterogeneous knowledge and the second path employs a dynamic graph attention network to capture the effects of task migration. The two paths are connected through a cross multi-head attention mechanism.
    \item Combining with continual learning, we design a two-stage training scheme for FT-MoE, consisting of offline training and online tuning. Offline training aims to optimize parameters. Online tuning aims to dynamically add or remove parameters to adapt to dynamic environments.
    \item We construct a new dataset for fault detection and classification. Our dataset includes more fine-grained host resource features and scheduling decisions, making it better reflect the dynamic environment than existing ones.
\end{itemize}

\section{Related Work}
\subsection{Fault Tolerant Computing Methods}
Most contemporary state-of-the-art FT computing methods employ some form of specialized algorithms or machine learning models. 
Specialized algorithms methods usually rely on predefined models to achieve FT tasks. 
DFTM~\cite{SIVAGAMI201935} detects faults by monitoring the load and network traffic of virtual machines (VM). The faults are classified as single VM faults and multi-node faults. 
PCFT~\cite{7469864} predicts in advance which physical machines are likely to fail based on CPU temperature. 
ECLB~\cite{article} technique uses Bayesian methods and neural networks to classify host machines into different categories.
CMODLB~\cite{Negi2021CMODLBAE} uses k-means and deep learning to cluster nodes and identify overloaded hosts.
In a recent work, PreGAN~\cite{9796778} uses gated recurrent units and graph attention networks to construct a deep learning model for fault prediction.
PreGAN+~\cite{10310127} combines the transformer structure, further improving the accuracy of fault prediction.
However, existing methods cannot handle heterogeneous fault knowledge, using the same model for all fault types and ignoring knowledge diversity. Most are offline trained, cannot adapt after deployment and perform worse with changing system conditions. In our experiments, we compare FT-MoE against the state-of-the-art baselines DFTM, ECLB, PCFT, CMODLB, PreGAN and PreGAN+.

\subsection{Mixture of Experts Model}

Mixture of experts~\cite{dai-etal-2024-deepseekmoe} is a machine learning method based on multiple expert models. Its core idea is to select suitable models for different inputs. 
Compared to traditional single models, MoE uses multiple experts to handle different types of faults, which helps to improve overall detection accuracy.
Due to the flexibility and efficiency, many recent studies have attempted to apply MoE to handle time-series data.
MoLE~\cite{DBLP:conf/aistats/Ni0WF24} trains multiple linear-centric models and a router model that weighs and mixes their outputs to predict time series.
MEO~\cite{he-etal-2023-merging} proposes combining multiple experts into a single shared expert to significantly improve the computational efficiency of the MoE while maintaining performance.
Time-MoE~\cite{DBLP:conf/iclr/ShiWNLYWJ25} combines a series of decoder-only transformer models with MoE, supporting flexible prediction ranges with different input context lengths. 
However, most existing MoE methods rely on statically trained architecture, which limits their adaptability to changing input distributions and dynamic environments.





\section{Dynamic Fault Detection and Recognizing Algorithm}
\begin{figure*}[ht]
\includegraphics[width=1\textwidth]{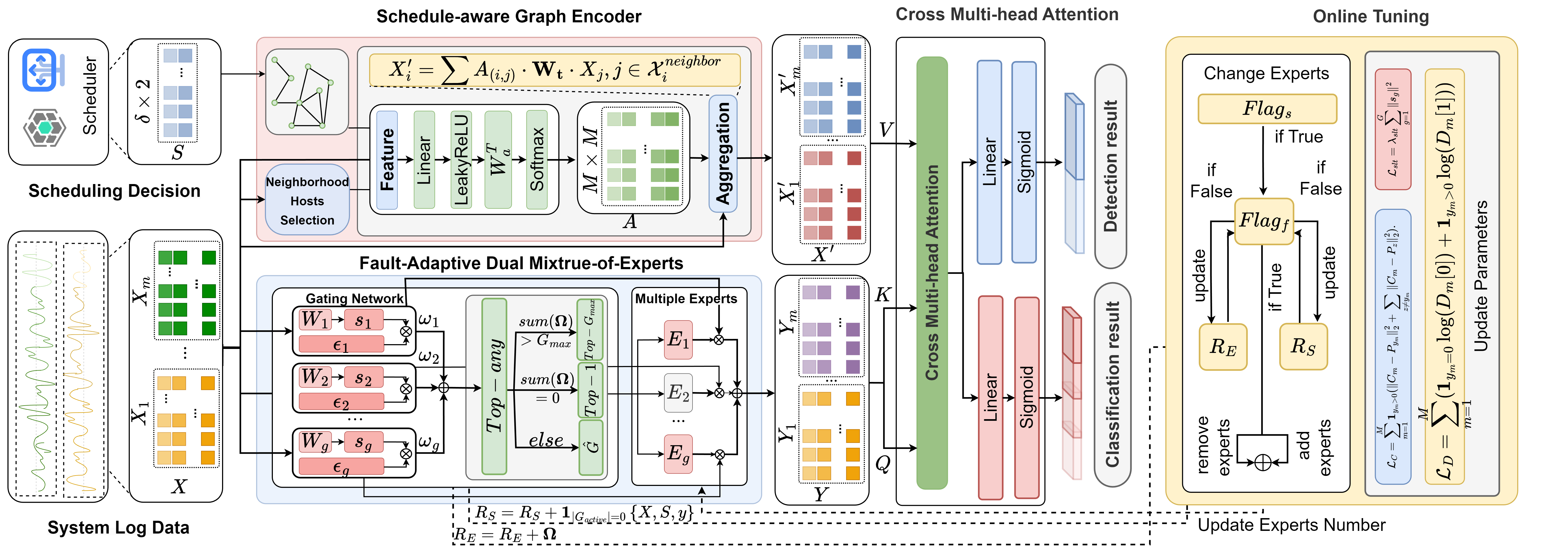}
    \caption{FT-MoE, including the schedule-aware graph encoder, the fault-adaptive mixture-of-experts layer, the cross multi-head attention layer, the offline training module and the online tuning module.}
    \label{fig:Model}
\end{figure*}

We propose FT-MoE, a sustainable-learning mixture-of-experts framework for fault-tolerant computing, aiming to improve fault prediction performance. As shown in Fig.~\ref{fig:Model}, FT-MoE consists of three components: a schedule-aware graph encoder to capture task propagation across hosts, a fault-adaptive MoE layer to handle fault heterogeneity and a cross multi-head attention layer to fuse heterogeneous features with scheduling information. In addition, we adopt a two-stage training strategy, including offline optimization for diverse fault scenarios and online tuning for continual adaptation to dynamic environments.

\subsection{FT-MoE System Model and Problem Statement}
In this work, we consider a standard heterogeneous environment, where all computational nodes are interconnected through a Local Area Network (LAN)~\footnote{We provide more details for system model in appendix 1}. 

\noindent\textbf{Definition 1.} (Host Set and Feature Matrix)
We assume that there are $M$ hosts in the environment, denoted as $ H = \{ h_1, \dots, h_m \} $, where $m \in M$. Each host has $ N $ resource utilization features. The feature vector of the host at interval $ t $ is represented as $X_t \in \mathbf{R}^{M \times N}$. 

\noindent\textbf{Definition 2.} (Task Scheduling Matrix) 
At the beginning of each interval $t$, a set of new tasks arrives at the system, and the scheduler generates a scheduling decision simultaneously. We assume that there are $\delta$ tasks running simultaneously in the system. The scheduling decision at the interval $t$, denoted as $S_t \in \mathbf{R}^{\delta \times 2}$, shows the task migration from the source host to the target host. Each migration creates an edge between two different hosts.





\subsection{Schedule-aware Graph Encoder}
We design the schedule-aware graph encoder to model the propagation dynamics of task scheduling behaviors across hosts. 
Specifically, the feature vector of $m$-th host is denoted as $ X_m \in \mathbf{R}^{N} $. The hosts that are connected to the $m$-th host in the scheduling decision $S$ are called the neighborhood hosts of the $m$-th host. The set of neighborhood hosts' indices for the $m$-th host is denoted as $ \mathcal{H}_m^{neighbor} $. Following previous study~\cite{DBLP:conf/iclr/Brody0Y22}, we calculate attention weights $A$~\footnote{We provide more details for calculating the attention weights $A$ in appendix 2}. $A_{(m, m')}$ represents the attention weight of the edge $(m,m')$. 
Then, we aggregate the attention weights of all neighborhood hosts to get the updated feature vector of the $i$-th host $X'_i$. The aggregation process can be represented as follows:
\begin{equation}
X'_i = \sum_{j \in \mathcal{H}_i^{neighbor}} A_{(i, j)} \cdot \mathbf{W_t} \cdot X_j,
\end{equation}
where $\mathbf{W_t}$ is trainable parameter. 
$X'$ denotes the set of updated feature vector, $X'_i \in X'$.

\subsection{Fault-Adaptive Mixture-of-Experts}
To handle the heterogeneity of FT tasks, we design a fault-adaptive mixture-of-experts layer. It consists of multiple experts that capture diverse fault patterns and a gating network that selects the most relevant experts based on the input.
\subsubsection{Experts-Adaptation Gating}

Traditional MoE uses a fixed number of active experts, but this often leads to inefficiency in practice~\cite{DBLP:journals/tmlr/MuqeethLR24}. To solve this, we introduce Experts-Adaptation Gating (EAGate), which automatically chooses and adjusts the number of experts based on input features for fault prediction. Next, we describe the specific process of the EAGate. EAGate first calculates the cosine similarity between the input and each expert. These similarity scores are then used to decide which experts should be activated. Let $ E_g $  denote the $g$-th expert of the MoE. Let $ W_g $ denote the $g$-th expert's representation matrix of the MoE, which is trainable in EAGate. $g \in G$ and $G$ is the number of experts in an MoE model. Hence, the similarity $s_g $ between $X$ and expert $ E_g $ is represented as:

\begin{equation}
s_g  = \frac{\langle X, W_g \rangle}{\|X\| \|W_g\|} 
\end{equation}
where \( \langle x_1, x_2 \rangle \) function denotes the dot product between \(x_1\) and \(x_2\) and \( \| \cdot \| \) is the L2 norms function. By normalizing \(X\) and \(W_g\), we obtain all standardized similarity scores which are denoted as $\mathbf{S} = \{ s_g \}$.

Then, we apply the sigmoid function to map the similarity score to [0,1] to represent the activation probability of experts, that is $\hat{\mathbf{S}} = \{\hat{s}_g\}$. We define the activation threshold vector as \(\epsilon \in \mathbf{R}^g\) and they are trainable parameters. \(\epsilon_g\) represents the $g$-th expert's activation threshold. We apply the sigmoid function to map the activation threshold to [0,1], that is $\hat{\epsilon}$.
By comparing $\hat{s}_g$ with the activation threshold \(\hat{\epsilon}_g \), we can determine whether $g$-th expert $E_g$ should be activated during a particular computation round. Let $\omega_g$ denote the activation state of expert $E_g$ and $\mathbf{\Omega} = \{\omega_g\}$ denote the activation state of all experts, its formula is represented as: 
\begin{equation}
\omega_g = \text{sign}(\hat{s}_g - \hat{\epsilon}_g),
\end{equation}
where sign($x$) function is used to convert $x$ to 0 or 1. If \( x > 0 \), then sign($x$) =1; else, sign($x$) =0. Thus, if \( \omega_g = 1 \), it indicates that expert $E_g$ can be activated; otherwise, it can not be activated.
We set the activatable expert indices as $\hat{G}$ and the activatable expert as $\hat{E}_{\hat{g}}$, where $\hat{g} \in \hat{G}$. We set the activatable experts as $\hat{E} = \{ \hat{E}_{\hat{g}} \}$.

Then, to enhance the model's adaptability and flexibility, we present a Top-any function to select a set of experts to activate, which enables each input to autonomously determine the number of experts to activate. The output of the Top-any function is a set of experts' indices. We set the maximum number of selectable active experts for one input as $G_{max}$. Specifically, if sum$(\mathbf{\Omega}) > G_{max}$, we select the $G_{max}$ experts with the highest similarity scores to be activated, experts' indices are denoted as Top-$G_{max}(\hat{s})$; if sum$(\mathbf{\Omega}) = 0$, we select an expert with the highest similarity scores to be activated, expert's index is denoted as Top-$1(\hat{s})$; else, we select the set of activatable experts $\hat{E}$, experts' indices are $\hat{G}$. Its formula is represented as:
\begin{equation}
\text{Top-any} = 
\begin{cases} 
\text{Top-}G_{max}(\hat{s}) & \text{if } \text{sum}(\mathbf{\Omega}) > G_{max}, \\
\text{Top-}1(\hat{s}) & \text{if } \text{sum}(\mathbf{\Omega}) = 0, \\
\hat{G} & \text{else}.
\end{cases}
\end{equation}
Based on Top-any, we can obtain experts selected to be activated, whose indices are denoted as $G_{active}$.

\subsubsection{Multiple Experts Module} 
The multiple experts module of an MoE consists of multiple sparse activation experts. To avoid model learning bias caused by unusually strong or weak outputs from some experts, we adopt a weighted sum approach to integrate the outputs of all activation experts~\cite{Sakib2024EnsembleDL}. Let $G_{active}$ denote the set of experts' indices selected to be activated. The output of the multiple experts module is denoted as $Y$, which is calculated as follows:
\begin{equation}
Y = \frac{1}{|G_{active}|}\sum_{g \in G_{active}}E_g s_g,
\end{equation}
where $|G_{active}|$ denotes the number of activation experts' indices in $G_{active}$.

\subsection{Cross Multi-head Attention Module} 
The cross multi-head attention (CMHA) module is used to model the relationship between the task scheduling behavior and the heterogeneous resource knowledge. 
Specifically, the attention is represented as:
\begin{equation}
\text{Attention}(\mathbf{Q}, \mathbf{K}, \mathbf{V}) = \text{softmax}\left( \frac{\mathbf{Q} \mathbf{K}^T}{\sqrt{d_k}} \right) \mathbf{V},
\end{equation}
where \(\mathbf{Q}\) is the query, \(\mathbf{K}\) is the key and \(\mathbf{V}\) is the value. \(d_k\) is the dimension of \(\mathbf{K}\). CMHA computes attention in parallel using multiple sets of \(\mathbf{Q}\), \(\mathbf{K}\) and \(\mathbf{V}\). Each set has its own weight matrix, is called a $head$ and captures different input aspects. $\beta$ is the number of heads. $head_a$ is the result of the computation for $a$-th $head$ and $a \in A$, represented as:
\begin{equation}
head_a = \text{Attention}(\mathbf{Q}W^Q_\beta, \mathbf{K}W^K_\beta, \mathbf{V}W^V_\beta),
\end{equation}
\begin{equation}
\text{MultiHead}(\mathbf{Q}, \mathbf{K}, \mathbf{V}) = [head_1; \dots; head_\beta] W_O,
\end{equation}
where $W^Q_\beta$, $W^K_\beta$, $W^V_\beta$ and $W_O$ are all trainable parameters. 
We set $Q = Y$ and $K = V = X'$. Let each host’s features actively search the graph to find which scheduling relationships affect them most, and use this to make better fault predictions. $O$ represents the context vector obtained from CMHA, which is represented as:
\begin{equation}
O = \text{MultiHead}(Y, X', X'),
\end{equation}

To adjust the feature dimensions and scale the output, we use two separate feed-forward networks: one for fault detection and one for fault classification. For fault detection, we apply a linear layer Linear$^D$ followed by a softmax to produce probabilities. For fault classification, we use another linear layer Linear$^C$ followed by a sigmoid. These two FFNs are referred to as FFN$^D$ and FFN$^C$, respectively. FFN$^D$ and FFN$^C$ are represented as:
\begin{equation}
\text{FFN}^{D} = \text{Softmax}(\text{Linear}^D(O)).
\end{equation}
\begin{equation}
\text{FFN}^{C} = \text{Sigmoid}(\text{Linear}^C(O)).
\end{equation}

Finally, FFN$^D$ and FFN$^C$ output the fault detection result $D \in \mathbf{R}^{M \times 2}$ and fault classification result $C \in \mathbf{R}^{M \times Q}$, respectively. Here, $D$ gives the fault prediction for all $M$ hosts, where $D_m$ represents the prediction for the $m$-th host. A fault is predicted in host $m$ if $D_m[1] > D_m[0]$.
The classification output $C$ gives each host’s fault feature vector, where $Q$ is the feature dimension and $C_m$ is the vector for host $m$.
We assume there are $Z$ fault types, with each type $z \in Z$ represented by a prototype vector $P_z$ in a matrix $P \in \mathbf{R}^{Z \times Q}$. These prototype vectors are trainable and initialized randomly in the range [0, 1].
To classify a fault, we compute the Euclidean distance between $C_m$ and each $P_z$, and select the fault type whose prototype is closest to $C_m$.

\subsection{Offline Training and Online Tuning}

In this section, we describe the offline training and the online fine-tuning. The offline training describes how to train the model with datasets acquired in a real environment. The online fine-tuning describes how to fine-tune the model based on real-time data when running the simulation platform.

\subsubsection{Offline Training}
In this section, we describe the offline training process for the FT-MoE to detect and classify faults. Datasets of FT computing consist of time-series data \( X \), scheduling decisions \( S \) and fault type labels \( y \). \( y_m \) represents $m$-th host's fault class, where \( y_m = 0\) denotes no fault. For the fault detection, we use cross-entropy loss as the loss function and sum the loss over all hosts to get the final loss, its formula is represented as:

\begin{equation}
\mathcal{L}_{D} = \sum_{m=1}^{M} (\mathbf{1}_{y_m = 0}\log(D_m[0]) + \mathbf{1}_
{y_m > 0}\log(D_m[1])) \label{L1},
\end{equation}
where \(\mathbf{1_{ \text{cond} }}\) is an indicator function, represented as:

\begin{equation}
\mathbf{1_{\text{cond}}} = 
\begin{cases}
1, & \text{if cond is true}, \\
0, & \text{if cond is false},
\end{cases}
\end{equation}

\noindent where cond denotes the condition.

For the fault classification, we use the triplet loss as the loss function~\cite{10.1007/978-3-030-01261-8_28}. We calculate the euclidean distance between the model-generated fault type prediction for $m$-th host \( C_m \) and each prototype vector in \( P \). Then the euclidean distance between \( C_m \) and the real category prototype vector $P_{y_m}$ is minimized while increasing the euclidean distance between \( C_m \) and the incorrect $z$-th class prototype vector $P_z$. Therefore, the fault classification loss $\mathcal{L}_{C}$ for all hosts with system faults is represented as:

\begin{equation}
\mathcal{L}_{C} = \sum_{m=1}^{M} \mathbf{1}_{y_m > 0} (\left\| C_m - P_{y_m} \right\|_2^2 + \sum_{z \neq y_m} \left\| C_m - P_z \right\|_2^2).\label{L2}
\end{equation}


Besides, we also introduce the expert selection loss, ensuring balanced activation frequencies for each expert~\cite{yang-etal-2024-xmoe}. Specifically, the similarity scores of experts are regularized using the L2 norms function. Let \(\lambda_{slt}\) be the regularization weight and $\mathcal{L}_{slt} $ denote the expert selection loss. its formula is expressed as:

\begin{equation}
  \mathcal{L}_{slt} = \lambda_{slt} \sum_{g=1}^{G} \| s_g \|^2 \label{Lr}
\end{equation}

Thus, the final loss function $\mathcal{L}_{final}$ combines $\mathcal{L}_{D}$, $\mathcal{L}_{C}$ and $\mathcal{L}_{slt}$, which is represented as:

\begin{equation}
  \mathcal{L}_{final} = \alpha_1 \mathcal{L}_{D} + \alpha_2 \mathcal{L}_{C} + \alpha_3 \mathcal{L}_{slt}\label{Lf}
\end{equation} 

\noindent where \(\alpha_1\), \(\alpha_2\) and \(\alpha_3\) are the weight parameters.

\subsubsection{Performance-aware Online Tuning}

To adapt to dynamic service environments, we design a performance-aware online tuning strategy for FT-MoE. It fine-tunes the fault-adaptive MoE using the host features of the previous interval $X_{t-1}$, scheduling decisions of the previous interval $S_{t-1}$ and fault labels of the current interval $y_t$. For simplicity, we denote the input as $\{X, S, y\}$. We still use $L_{final}$ to update the parameters. During online tuning, all components of FT-MoE are frozen except for the MoE module. Its key idea is to remove experts with low contribution and add new ones to handle previously unseen data, enabling efficient continual learning under changing conditions.
 
In detail, we record the active state of all experts to keep track of which experts are frequently activated, denoted as \( R_E \in \mathbf{R}^{G} \). \( R_{E,g}\) represents the times the $g$-th expert is activated during this inference. Its formula is represented as:
\begin{equation}
R_E = R_E + \mathbf{\Omega}
\end{equation}
Furthermore, we can also record the activated experts number of each input, denoted as \( R_S \). So we can find inputs that did not activate any experts. Its formula is represented as:
\begin{equation}
R_S = R_S +  \mathbf{1}_{|G_{active}|=0}\,{\{X,S,y\}}
\end{equation}
We use $flag_s$ and $flag_f$ to determine when to start and stop the routing recording. We initialize $flag_s = 1$ and $flag_f = 0$ and set an epoch threshold $I$. If the current epoch number is a multiple of $I$, we set $flag_f = 1$. Based on these history states, we can fine-tune fault-adaptive MoE according to the following strategy: 

(1) If \( R_{E,g} = 0 \), it means that the $g$-th expert has not been activated by any input during the training process. The $g$-th expert will be removed from the network to reduce the computational resources and improve the training efficiency.

(2) If \( R_S = NULL \), it means that there are inputs that do not have suitable experts and the system will automatically add a new expert. In addition, the new expert threshold $ \hat{\epsilon}_{new} $ is initially set to 0 to ensure that the inactive input can activate the newly added expert. 

After adding experts and deleting experts, we set $flag_f = 0$ and restart the routing recording.


\subsection{Our Dataset}
We build a new dataset to enable accurate and adaptable fault prediction in dynamic environments. It supports both supervised learning and continual adaptation. Compared to datasets like Google Cluster Trace, Alibaba Cluster Trace v2021, and Bitbrains GWA-T-12 VM Trace, ours uniquely includes full resource metrics, detailed scheduling logs, and host-level fault labels~\footnote{We provide more details for public datasets and comparison between them and ours in appendix 3 and 4}. 

To simulate realistic and dynamic workloads, we design a task generator using the Bitbrains GWA-T-12 dataset. We select the 'rnd' subset with 500 VMs and filter those whose CPU usage at the 10th interval is between 500 and 3000 MIPS to build a candidate VM pool.
In each simulation step, we randomly generate containers based on a Gaussian distribution. Each container uses a VM trace from the pool, and its full resource usage over time is used to simulate container behavior.
We use the COSCO framework~\cite{9448450} to simulate our fault-tolerant system with a 16 node distributed cluster. Eight nodes have 4 GB of RAM, and the other eight have 8 GB. The default scheduler, Gradient Based Optimization task scheduling strategy (GOBI)~\cite{9448450}, assigns containers based on host load and task demands. During execution, we log scheduling decisions and resource usage. 
Based on this simulator, we obtain host resource status data in various service environments as the training dataset.
Anomaly Detection Engine for Linux Logs (ADE) tool~\cite{AGYEMANG2024e02386} is used to generate ground-truth labels for our dataset. Fault types consist of CPU fault, RAM fault and disc fault~\footnote{We provide more details on how to determine the type of fault in appendix 5}. 
Our dataset contains 115,010 no fault samples, 8,001 CPU faults, 22,771 RAM faults and 14,218 Disk faults.



\section{Experiment}
\begin{table*}[t]
    \centering
    \setlength{\tabcolsep}{2.8pt} 

    \begin{tabular*}{\textwidth}{@{\extracolsep{\fill}}lccccccccc} 
        \toprule
        Method      & \multicolumn{4}{c}{Detection} & \multicolumn{2}{c}{Classification} & \\
        \cline{2-5} \cline{6-7}
                    & Accuracy         & Precision        & Recall           & F1 Score         & HR            
                    & NDCG          
                    \\
        \midrule
        DFTM        & 0.8210 $\pm$ 0.0200 & 0.7100 $\pm$ 0.0700 & 0.8100 $\pm$ 0.0200 & 0.7560 $\pm$ 0.0350 & 0.4800 $\pm$ 0.0200 & 0.4400 $\pm$ 0.0020 \\
        ECLB        & 0.8800 $\pm$ 0.0180 & 0.7350 $\pm$ 0.0650 & 0.8500 $\pm$ 0.0200 & 0.7880 $\pm$ 0.0400 & 0.5050 $\pm$ 0.0100 & 0.4700 $\pm$ 0.0020 \\
        PCFT        & 0.8500 $\pm$ 0.0120 & 0.7600 $\pm$ 0.0500 & 0.8700 $\pm$ 0.0180 & 0.8110 $\pm$ 0.0300 & 0.5650 $\pm$ 0.0090 & 0.5200 $\pm$ 0.0020 \\
        CMODLB      & 0.8700 $\pm$ 0.0100 & 0.7750 $\pm$ 0.0300 & 0.8750 $\pm$ 0.0150 & 0.8225 $\pm$ 0.0250 & 0.5950 $\pm$ 0.0030 & 0.5300 $\pm$ 0.0030 \\
        PreGAN      & 0.9000 $\pm$ 0.0080 & 0.8200 $\pm$ 0.0220 & 0.8950 $\pm$ 0.0120 & 0.8555 $\pm$ 0.0200 & 0.6050 $\pm$ 0.0060 & 0.5650 $\pm$ 0.0050 \\
        PreGAN+     & \textbf{0.9060 $\pm$ 0.0070} & \textbf{0.8290 $\pm$ 0.0060} & 0.9180 $\pm$ 0.0080 & 0.8710 $\pm$ 0.0180 & 0.6250 $\pm$ 0.0030 & 0.5800 $\pm$ 0.0040 \\
        \hline
        \textbf{FT-MoE}     & 0.9053 $\pm$ 0.0028 & 0.8273 $\pm$ 0.0079 & \textbf{0.9322 $\pm$ 0.0094} & \textbf{0.8766 $\pm$ 0.0061} & \textbf{0.6496 $\pm$ 0.0055} & \textbf{0.6021 $\pm$ 0.0027} \\
        \bottomrule
    \end{tabular*}
    \caption{Overall comparison of methods across the fault detection and the fault classification.}
    \label{tab:full_results}
\end{table*}

\subsection{Experimental Setting}
For FT-MoE, we set the dimensionality of the prototype vector \(Q\) at 8. For gradual updating of prototype vectors $P$, we adopt a hybrid learning prototype vector updating scheme~\footnote{We provide more details for hybrid-learning prototype vectors updating scheme in appendix 6}. The number of experts in an MoE model $G$ is numbered 12 and the maximum number of selectable active experts for one input $G_{max}$ is numbered 8. For the offline training FT-MoE model, we use the AdamW optimizer to train FT-MoE offline~\cite{9014697}, with an initial learning rate of 0.001 and decays by 0.1 every 10 epochs. We set the weights of the loss function \(L_{fin}\): \(\alpha_1\), \(\alpha_2\) and \(\alpha_3\) to 0.35, 0.5 and 0.15, respectively. The update weight \( \eta \) is set to 0.9. Note that we run all experiments for 100 scheduling intervals and average over 5 runs as final performance results. We use the COSCO framework as the service simulator for our experimental.

\subsection{Evaluation Metrics}

\textbf{Fault Prediction}. Our framework includes two tasks of the fault detection and the fault classification. They have different evaluation metrics: (1) For the fault detection, we use precision, recall and F1 score as prior work~\cite{Anwar2025RobustFD}. (2) For the fault classification, we use two metrics, that is HitRate (HR)~\cite{10.14778/3514061.3514067} and Normalized Discounted Cumulative Gain (NDCG)~\cite{10.1145/582415.582418}~\footnote{We provide details for experimental metrics in appendix 7}. 

\textbf{Fault-Tolerant Scheduling}. 
Quality of Service (QoS) scores are some metrics for evaluating the quality of user service. Here, we adopt the CPU utilization, the RAM utilization, the energy consumption and the response time metrics to measure the system performance~\footnote{We provide additional QoS results in appendix 8}.

\subsection{Experimental Results}
\begin{figure}[ht]
    \centering
    \subfigure[CPU Utilization]{
        \includegraphics[width=0.484\linewidth]{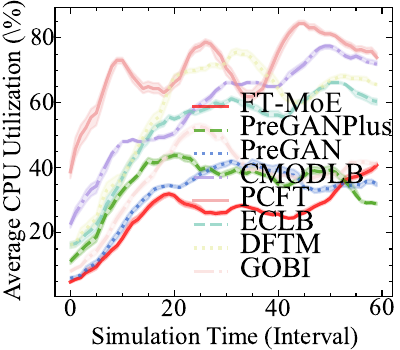}\label{fig:CPUS}}
    \hfill 
    \subfigure[RAM Utilization]{
        \includegraphics[width=0.484\linewidth]{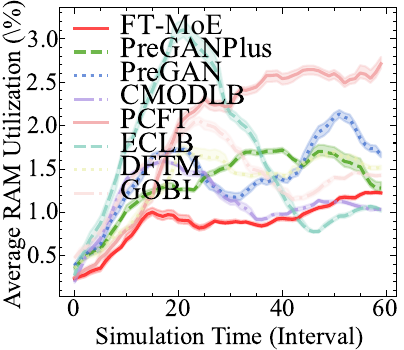}\label{fig:RAM}}
    
    \subfigure[Energy Consumption]{
        \includegraphics[width=0.484\linewidth]{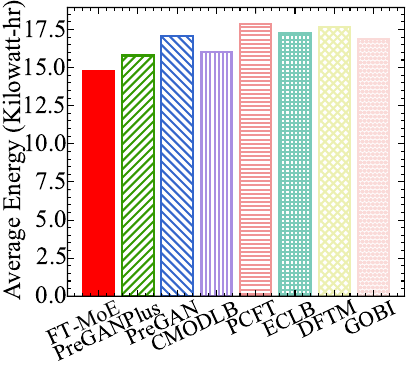}\label{fig:Energy}}
    \hfill
    \subfigure[Response Time]{
        \includegraphics[width=0.484\linewidth]{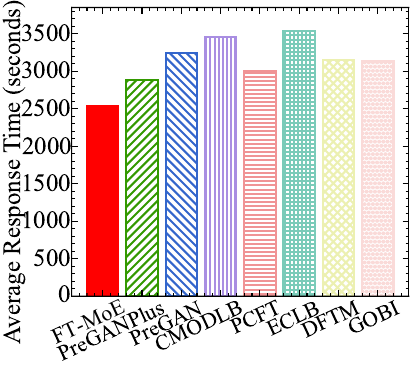}\label{fig:response}}
        
    \caption{QoS performance of FT-MoE with baselines.}
    \label{fig:QoS}
\end{figure}
To evaluate the performance of FT-MOE, we conducted comparative experiments with the latest baselines: DFTM~\cite{SIVAGAMI201935}, ECLB~\cite{article}, PCFT~\cite{7469864}, CMODLB~\cite{Negi2021CMODLBAE}, PreGAN~\cite{9796778} and PreGAN+~\cite{10310127}. 
First, we evaluate the performance of the fault detection and the fault classification, which have different evaluation metrics.
Then, to verify the effect of FT-MoE on avoiding system overload, we further evaluate the task migration performance after a successful fault detection. 

Following ~\cite{10310127}, we use the Generative Adversarial Networks (GAN) to decide tasks scheduling decisions based on predicted faults, which contains a generator and a discriminator network. We analyze the QoS scores after task scheduling decisions based on the prediction results of our model.
\textbf{Comparison of Fault Prediction Methods}.
The experimental results are shown in Table~\ref{tab:full_results}. 
(1) In the fault detection, FT-MoE outperforms other baseline methods in recall and F1 score respectively. FT-MoE achieves 0.9041 in accuracy, which outperforms most other baseline methods. This is due to FT-MoE capturing high-dimension features of input $X_t$ and $S_t$ better than other methods. 
(2) In the fault classification, FT-MoE's HR and NDCG showed an improvement of 3.936\% and 3.81\% respectively, over the best baseline method. This is due to FT-MoE considering knowledge heterogeneity and using different sets of parameters for inputs with different types of feature. 
While baseline models use identical parameters.

\textbf{System Scheduling QoS Breakdown}.
Fig.~\ref{fig:QoS} compares some QoS scores of all models, including the original GOBI without any fault-tolerant method. For all the fault prediction models, we use the same scheduling decision generator~\cite{10310127} to generate scheduling decisions based on the fault prediction results. 
(1) Fig.~\ref{fig:QoS}(a) and Fig.~\ref{fig:QoS}(b) indicates that FT-MoE has lower average CPU and RAM utilization compared to the baselines. This is due to the higher fault detection and fault classification accuracy of our model, which helps the generator generate better scheduling decisions. 
(2) Due to the lower average CPU and RAM utilization, FT-MoE also has the lowest average energy consumption of 14.1641 KW-hr, followed by PreGAN+ with 16.4324 KW-hr, as Fig.~\ref{fig:QoS}(c) shows. 
(3) Fig.~\ref{fig:QoS}(d) shows a comparison of FT-MoE and the baselines with the response time. It shows that FT-MoE has the minimum response time. This is due to FT-MoE, which has a faster inference speed compared to the baseline methods. 

\subsection{Ablation Study}

In this section, we conduct ablation studies to analyze the effects of different components in FT-MoE~\footnote{We provide more experiment results in appendix 9}.
\subsubsection{Analysis of EAGate}
\begin{figure}[h]
    \centering
    \subfigure[Wo/EAGate]{
    \includegraphics[width=4cm]{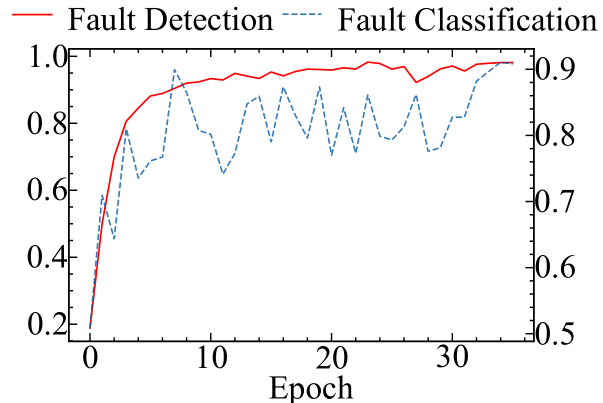}}
    \subfigure[W/EAGate]{
    \includegraphics[width=4cm]{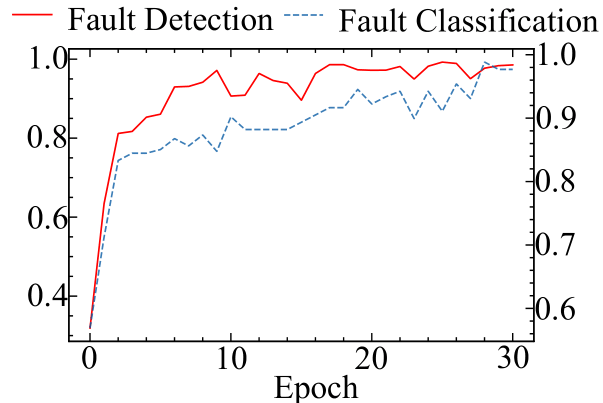}}
    \caption{Training curves for the fault detection and classification.}
    \label{fig:trainaccuary}
\end{figure}

In this section, we evaluate the impact of the proposed EAGate. Fig.~\ref{fig:trainaccuary} presents the accuracy trends for both the fault detection and the fault classification during training, comparing FT-MoE models with the EAGate (W/EAGate) and without it (Wo/EAGate). In the Wo/EAGate configuration, the EAGate is replaced by a static gating. 
We could observe that: (1) The fault classification accuracy of the Wo/EAGate fluctuates drastically and the optimal accuracy only reaches 87.956\%. This is because static gating treats all diverse fault type inputs uniformly, ignoring their differences. (2) In contrast, the W/EAGate configuration exhibits a steady improvement in classification accuracy, ultimately reaching 90.479\%. Because the EAGate can adaptively select experts based on input features, enabling more fine-grained training and capturing heterogeneous knowledge of different types of fault.
In general, these results show that the EAGate plays a key role in stabilizing training and improving the model’s ability to extract meaningful features for accurate fault classification.

\subsubsection{Analysis of Online Tuning}
\begin{figure}[h]
    \centering
    \subfigure[Wo/OnlineTuning]{
    \includegraphics[width=4cm]{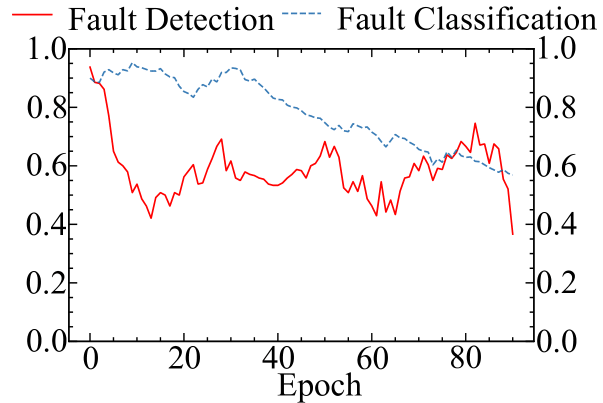}}\label{fig:Wotuning}
    \subfigure[W/OnlineTuning]{
    \includegraphics[width=4cm]{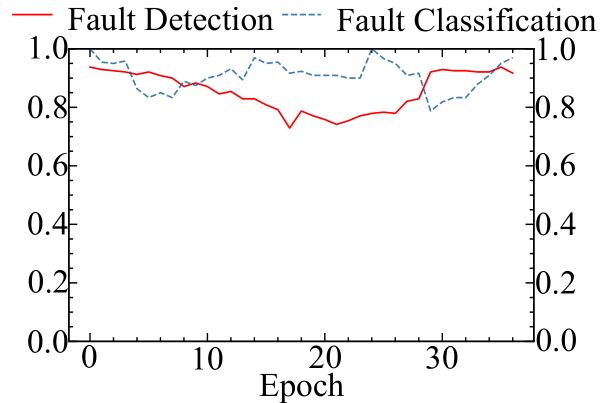}}\label{fig:Wtuning}
    \caption{Fine-tuning curves for the fault detection and classification.}
    \label{fig:finetune}
\end{figure}
In this section, our aim is to demonstrate the effects of online tuning. Fig.~\ref{fig:finetune} shows the changes in FT tasks accuracy during running in the edge environment of the FT-MoE with online tuning (W/OnlineTuning) and without it (Wo/OnlineTuning). Fig.~\ref{fig:finetune}(a) shows that after a period of running, the performance of Wo/OnlineTuning decreases greatly and finally reduces to 52.624\% for the fault detection and 57.223\% for the fault classification. This is because Wo/MoE uses unchanged parameters, so its ability to generalize decreases as new data comes in. This phenomenon is common in continual learning and is called plasticity loss~\cite{10.5555/3709347.3743870}. Fig.~\ref{fig:finetune}(b) shows the running results for W/OnlineTuning. Although there is also a performance loss for a while after the start of the fine-tuning, it then stabilizes. After that, the performance is maintained at a high level. This is because W/OnlineTuning has a stronger generalization ability compared to Wo/OnlineTuning.
\subsubsection{Analysis of CMHA}
\begin{figure}[h]
    \centering
    \subfigure[Wo/CMHA]{
    \includegraphics[width=4cm]{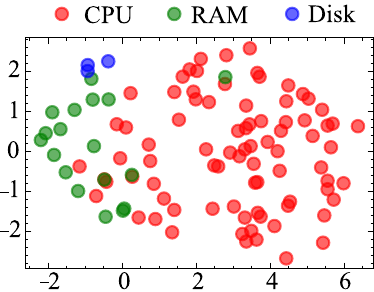}}\label{fig:WoCrossAttnC}
    \subfigure[W/CMHA]{
    \includegraphics[width=4cm]{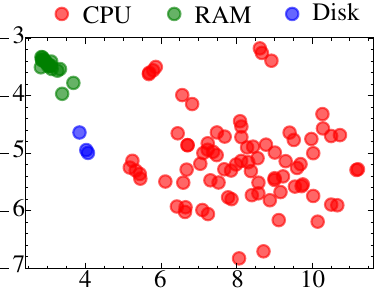}}\label{fig:WCrossAttnC}
    \caption{T-SNE plot for the fault classification using Wo/CMHA and W/CMHA.}
    \label{fig:CrossAttnC}
\end{figure}
In this section, our aim is to demonstrate the necessity of the CMHA module.
Fig.~\ref{fig:CrossAttnC} includes t-SNE visualizations of the predicted class prototypes for the fault classification. We observe that the embeddings produced by the model with CMHA (W/CMHA) are more compact and well separated by fault type. This indicates stronger intra-class consistency and higher inter-class discrimination. In contrast, the embeddings of the model without CMHA (Wo/CMHA) are more dispersed, leading to poorer classification performance. 
Additionally, W/CMHA also achieves significantly better alignment with the ground truth than Wo/CMHA, particularly in accurately localizing faults across different hosts and time periods. 
These results show that CMHA improves feature interaction and fault prediction by better associating task behavior with host status.

\section{Conclusion}
In this paper, we propose FT-MoE, a multitask learning framework for fault awareness and adaptive fault-tolerant learning. It leverages a MoE network with efficient gating to capture diverse fault patterns, and adopts a two-stage training strategy, including offline optimization and online continual tuning, to adapt to dynamic environments. Experiments on a simulated edge setting show that FT-MoE outperforms existing methods. Ablation studies confirm the effectiveness of each module in supporting multitask fault tolerance. 
In future work, our goal is to validate FT-MoE in real-world deployments and incorporate federated learning for privacy to assess its robustness and scalability.

\section{Acknowledgments}
This work was supported in part by the National Natural Science Foundation of China  (Nos. 62502101, 62462002) and partially supported by the Natural Science Foundation of Guangxi, China (Nos. 2025GXNSFAA069958, 2025GXNSFBA069394)

\bigskip

\bibliography{aaai2026}

\end{document}


\maketitle
\vspace{-30pt}
\section{System Model}
Within the IoT layer, various sensor and actuator devices generate data and initiate tasks in the form of lightweight container instances. These tasks are transmitted to the edge resource layer through gateway devices, which act as intermediaries between the physical and computational domains.
At the edge management layer, the task management is centralized under the control of an edge broker, which is responsible for critical operational decisions. These include task scheduling, lifecycle management, and preemptive migration of tasks between edge hosts to ensure load balancing and fault tolerance. To enhance the reliability of these operations, we deploy our proposed FTEdge model within this layer. FT-MoE serves as an intelligent fault prediction module that analyzes real-time resource features and task migration behaviors to proactively detect host faults and classify fault types. Its predictions help the edge broker make better scheduling decisions. 
The edge hosts layer, where actual task execution occurs, is composed of heterogeneous machines, each with varying computational capabilities and resource profiles.
\begin{table*}[t]
\centering
\caption{Comparison of Our Dataset with Popular Public Datasets}
\label{tab:dataset-comparison}
\begin{tabular}{|p{1.8cm}|p{6cm}|p{2cm}|p{4cm}|}
\hline
\textbf{Dataset Name} & \textbf{Monitored Resource Metrics} & \textbf{Scheduling Information} & \textbf{Failure Classification} \\
\hline
\textbf{Google Cluster Trace~\cite{clusterdata:Wilkes2020}} & CPU usage, RAM usage, disk usage & not available & No host-level failure types \\
\hline
\textbf{Alibaba Cluster Trace v2021~\cite{luo2021characterizing}} & CPU usage, RAM usage, disk read/write throughput, task DAG structure & not available & No host-level failure types \\
\hline
\textbf{Bitbrains GWA-T-12 VM Trace~\cite{10.1002/ett.4933}} & VM-level CPU usage, RAM usage, disk read/write throughput, network read/write throughput & not available & No host-level failure types \\
\hline
\textbf{Our Dataset} & \textbf{CPU usage, RAM usage, RAM read/write throughput, Disk usage, Disk read/write throughput} & \textbf{available} & \textbf{4-class host-level fault types: No fault, CPU fault, RAM fault, Disk fault} \\
\hline
\end{tabular}
\end{table*}

\begin{figure}[ht]
    \centering
    \includegraphics[width=8cm]{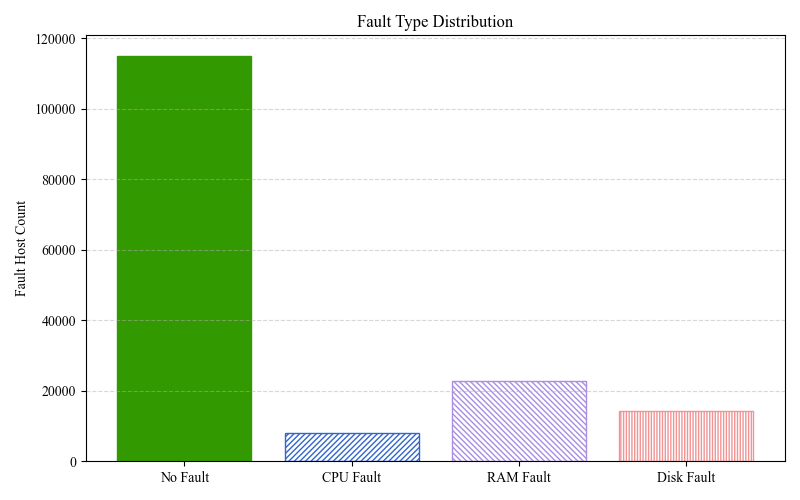}
    \caption{Fault Type Distribution of Our Dataset.}
    \label{fig:distribution}
\end{figure}

\section{Attention Weights}
Specifically, the feature vector of $m$-th host is denoted as $ X_m \in \mathbf{R}^{N} $. The hosts that are connected to the $m$-th host in the scheduling decision $S$ are called the neighborhood hosts of the $m$-th host. The neighborhood hosts' indices set of the $m$-th host is denoted as $ \mathcal{H}_m^{neighbor} $. We count how many distinct edges there are in the scheduling decision $S$ and $K$ denotes the number of distinct edges. Then, we count how many tasks migrate on each distinct edge and denote it as $F \in \mathbf{R}^{K \times 1}$. Any row of matrix $F$ consists of a pair data $(m,m')$, which represents a directed edge pointing from host $m$ to host $m'$, and an integer, which represents the number of tasks migrating on this edge. We denote the number of tasks migrated on the edge $(m,m')$ as $F[(m,m')]$. We use $Linear_f$ to adjust the vector dimensions of $F$.
First, we compute the attention weights $A$. $A_{(m, m')}$ represents the attention weight of the edge $(m,m')$. $e_{(m, m')}$ denotes the importance of the features of the neighborhood host $m'$ to the host $m$. The process can be represented as:
\begin{equation}
F' =   \mathbf{Linear_f} (F[(m, m')]) 
\end{equation}
\begin{equation}
e_{(m, m')} =  \mathbf{W_a}^\top \cdot \text{LeakyReLU}\left( \mathbf{W_e} (X_m + X_{m'} +  F') \right)
\end{equation}
where $\mathbf{W_a}$ and $\mathbf{W_e}$ are all trainable parameters and $\mathbf{W_e}$ are the parameters of a linear layer. 

Then, we convert $e_{(i,j)}$ to an all-positive attention score $A_{(i, j)}$, ensuring that the sum of the attention score for all neighborhood hosts of each host is 1. The process can be represented as:
\begin{equation}
A_{(i, j)} =  \frac{\exp\left(e_{(i, j)}\right)}{\sum_{k \in \mathcal{H}_i^{neighbor}} \exp\left(e_{(i, k)}\right)}
\end{equation}

\section{Public Datasets}
\textbf{Google Cluster Trace~\cite{clusterdata:Wilkes2020}}: The Google Cluster Trace is a large-scale dataset collected from a production Google data center. It includes task-level logs such as submission and termination events, along with resource usage such as CPU and memory.

\textbf{Alibaba Cluster Trace v2021~\cite{luo2021characterizing}}: The Alibaba Cluster Trace v2021 contains eight days of workload data from a real production environment, including both long-term services tasks and short batch tasks. The dataset provides the tasks information and the VMs resource usage. 

\textbf{Bitbrains GWA-T-12 VM Trace~\cite{10.1002/ett.4933}}: The Bitbrains GWA-T-12 dataset records the time series resource usage of VMs at 5-second intervals, including CPU, memory, disk, and network metrics. It is often used to simulate real workloads due to its high granularity. 

However, existing datasets do not include actual task scheduling or host-level fault labels, which limits their use in fault diagnosis or fine-grained scheduling research.

\section{Comparison Between Public Datasets And Our Proposed Dataset}

Compared to existing datasets such as Google Cluster Trace, Alibaba Cluster Trace v2021, and Bitbrains GWA-T-12 VM Trace, our dataset uniquely provides comprehensive resource metrics, detailed scheduling information and host-level fault classification, enabling more effective and fine-grained fault-aware computing research. 

The comparison between public datasets and our proposed dataset is summarized in Table~\ref{tab:dataset-comparison}
Our dataset contains a total of 160,000 host status records, including 115,010 no fault samples, 8,001 CPU faults, 22,771 RAM faults, and 14,218 Disk faults, as Fig~\ref{fig:distribution} shows.

\section{How To Determine The Type Of Fault}
Anomaly Detection Engine for Linux Logs (ADE) tool~\cite{AGYEMANG2024e02386} is used to generate ground-truth fault labels for our dataset. Fault types consist of CPU over-utilization (CPUO), abnormal disk utilization (ADU), memory leak (MEL) and abnormal memory allocation (AMA). These criteria are applied to simulation traces to generate a labeled dataset for fault-tolerant learning. 
CPUO is flagged when CPU usage remains above 95\%. ADU occurs when disk read or write throughput exceeds the 90th percentile of historical values or disk usage remains above 95\%. MEL is identified when RAM read or write throughput exceeds the 90th percentile of historical values, while AMA is detected when RAM usage remains above 95\%.
For simplicity, CPUO is classified as a CPU fault, ADU as a disc fault and MEL/AMA as a RAM fault.
\section{Hybrid-Learning Prototype Updating Scheme}
A step in the training process is to update the prototypes vectors $P$. The prototype vectors are randomly initialized at the beginning of training, which is bad at classifying fault classes. $\mathcal{L}_{C}$ adopts prototype vectors of fault, enabling the prototype vectors to be updated as model parameters. However, it causes performance inefficient by only using  $\mathcal{L}_{final}$, since it update slowly. Thus, we adopt a hybrid-learning prototype vectors updating scheme, which gradually updates the prototype vector by combining fault class information. It includes two stage updating processes of fast starting and steady enhancement. 
We first use fast starting. Fast starting aims to quickly learning the knowledge from time-series data. If $m$-th host's fault classification result $C_m$ belongs to the correct class $y_m$, we can update the prototype vector of that class $P_{y_m}$. We update $P_{y_m}$ by weighting and summing the $C_m$ and $P_{y_m}$. \( \eta \) is the update weight. In our setting, the update step size \(\eta\) of the prototype vectors is set to 0.9. Its formula is expressed as:
\begin{equation}
P_{y_i} = (1 - \eta) P_{y_i} + \eta C_i.
\label{update}
\end{equation}

If the model performance continues to fluctuate, then we use steady enhancement. Steady enhancement aims to conduct fine-tuning adjustments on the model to improve its accuracy and stability. We stop using eq.\eqref{update} and only use triplet loss $\mathcal{L}_{C}$ to adopts prototype vectors of fault.

\section{Evaluation Metrics}

\textbf{Fault Prediction}.  Our framework includes two tasks of the fault detection and the fault classification. They have different evaluation metrics: (1) For the fault detection, we use precision, recall and F1 score as prior work~\cite{Anwar2025RobustFD}. (2) For the fault classification, we use two metrics, that is HitRate (HR)~\cite{10.14778/3514061.3514067} and Normalized Discounted Cumulative Gain (NDCG)~\cite{10.1145/582415.582418}. HR measures how many true labeled dimensions are predicted by the model.
NDCG is a ranking-based metric originally used in recommendation systems. In FTEdge, the fault classification is performed by comparing the predicted prototype vector of the $m$-th host $C_m$ with a set of predefined fault type prototype vectors $\{ P_0, P_1, \dots, P_{z-1} \}$. The euclidean distance between $C_m$ and all predefined fault type prototype vectors are calculated to produce a ranked list of candidate fault types. We use the ranked list to calculate the NDCG.

\noindent \textbf{Fault-Tolerant Scheduling}. 
QoS (Quality of Service) is a set of technologies that allow users to obtain predictable service levels in terms of throughput, delay jitter, delay, and packet loss rate. QoS scores are some metrics for evaluating user service quality. Here, we adopt CPU utilization, RAM utilization, energy consumption, SLA violations, wait time and response time metrics to measures this system performance.

\section{Additional Results For System Scheduling QoS Breakdown}
\begin{figure}[ht]
    \centering
    \subfigure[SLA Violations]{
    \includegraphics[width=4cm]{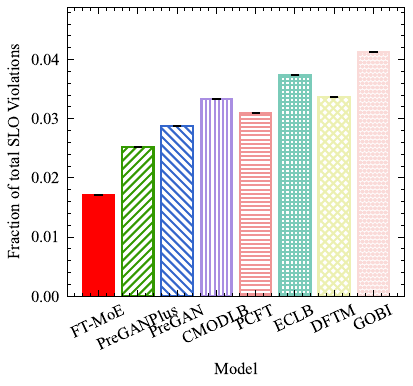}\label{fig:SLA}}
    \subfigure[Wait Time]{
    \includegraphics[width=4cm]{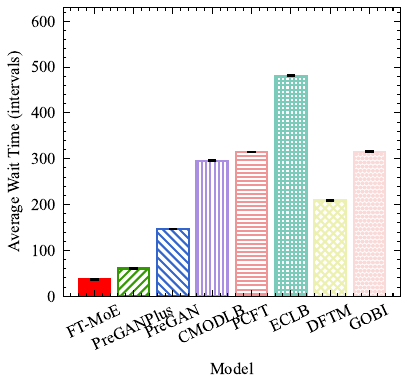}\label{fig:wait}}
    \subfigure[CPU Utilization]{
    \includegraphics[width=4cm]{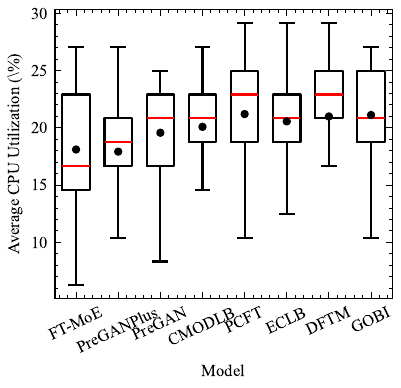}\label{fig:CPU}}
    \\
    \subfigure[RAM Utilization]{
    \includegraphics[width=4cm]{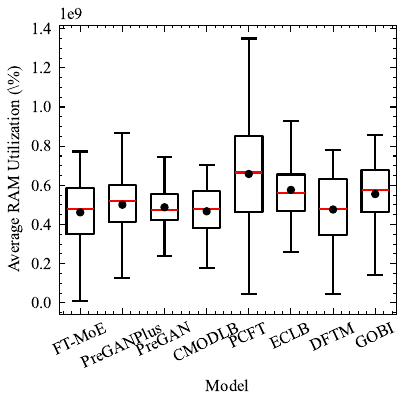}\label{fig:RAM}}
    \subfigure[Energy Consumption]{
    \includegraphics[width=4cm]{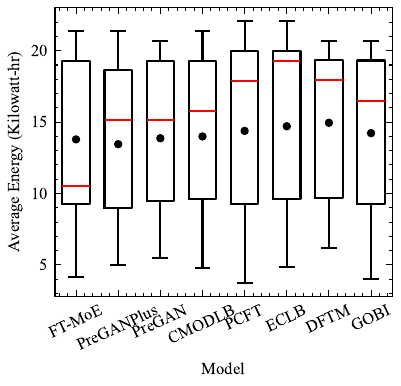}\label{fig:Energy}}
    \subfigure[Response Time]{
    \includegraphics[width=4cm]{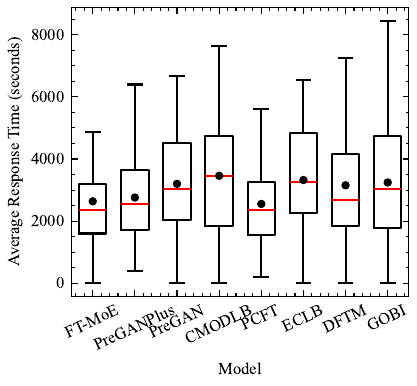}\label{fig:response}}
    \caption{Additional results for system scheduling QoS breakdown.}
    \label{fig:QoS}
\end{figure}
Fig.~\ref{fig:QoS} shows additional results for system scheduling QoS scores.
(1)Fig.~\ref{fig:CPU} and Fig.~\ref{fig:RAM} shows box plots of CPU utilization and RAM utilization. We can observe that FT-MoE has both the lowest CPU utilization and RAM utilization, which is consistent with the description in our paper.
(2)Fig.~\ref{fig:SLA} shows the bar chart of SLA violations and Fig.~\ref{fig:Energy} shows the box chart of energy consumption. Due to lower average CPU and RAM utilization, FT-MoE has the lowest SLO and energy consumption, as Fig.~\ref{fig:SLA} and Fig.~\ref{fig:Energy} shows. 
(3)Fig.~\ref{fig:wait} shows the bar chart of wait time which shows FT-MoE has the minimum wait time. Fig.~\ref{fig:response} is box chart of response time, it shows that FT-MoE has the minimum response time. This is due to FT-MoE has faster inference and result generation speed compared to the baseline methods.

\begin{figure}[ht]
    \centering
    \subfigure[Wo/CMHA]{
    \includegraphics[width=4cm]{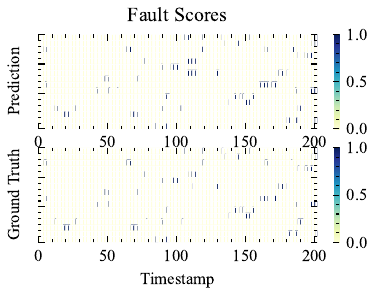}}\label{fig:WoCrossAttnD}
    \subfigure[W/CMHA]{
    \includegraphics[width=4cm]{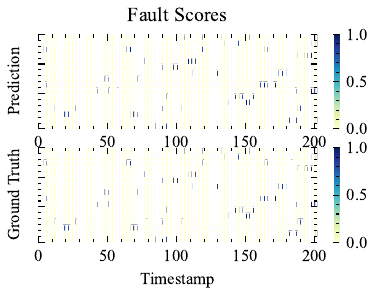}}\label{fig:WCrossAttnD}
    \caption{Fault detection using Wo/CMHA and W/CMHA with ground truth labels.}
    \label{fig:CrossAttnD}
\end{figure}
\section{Additional Results For Ablation Study}
\begin{table}
    \centering
    \small
    \setlength{\tabcolsep}{1.5pt} 
    \caption{Overall comparison of methods across the fault detection and the fault classification.}
    \begin{tabular*}{\textwidth}{@{\extracolsep{\fill}}lccccccccc} 
        \hline
        Method      & \multicolumn{4}{c}{Detection} & \multicolumn{2}{c}{Classification} & \\
        \cline{2-5} \cline{6-7}
                    & Accuracy         & Precision        & Recall           & F1 Score         & HR            
                    & NDCG          
                    \\
        \hline
        Wo/EAGate     & 0.8880 $\pm$ 0.0100 & 0.7980 $\pm$ 0.0080 & 0.9120 $\pm$ 0.0100 & 0.8510 $\pm$ 0.0120 & 0.5980 $\pm$ 0.0050 & 0.5600 $\pm$ 0.0030 \\
        Wo/OnlineTuning     & 0.8840 $\pm$ 0.0090 & 0.7950 $\pm$ 0.0070 & 0.9090 $\pm$ 0.0090 & 0.8480 $\pm$ 0.0110 & 0.5940 $\pm$ 0.0040 & 0.5570 $\pm$ 0.0030 \\
        Wo/CMHA     & 0.8710 $\pm$ 0.0110 & 0.7700 $\pm$ 0.0100 & 0.9100 $\pm$ 0.0120 & 0.8345 $\pm$ 0.0150 & 0.5750 $\pm$ 0.0060 & 0.5400 $\pm$ 0.0040 \\
        \hline
        \textbf{FT-MoE}     & \textbf{0.9053 $\pm$ 0.0028} & \textbf{0.8273 $\pm$ 0.0079} & \textbf{0.9322 $\pm$ 0.0094} & \textbf{0.8766 $\pm$ 0.0061} & \textbf{0.6496 $\pm$ 0.0055} & \textbf{0.6021 $\pm$ 0.0027} \\
        \bottomrule
    \end{tabular*}
    \label{tab:full_results}
\end{table}
Fig.~\ref{fig:CrossAttnD} presents the predicted and ground-truth fault labels on the test set, where the y-axis represents the host IDs and the x-axis denotes the time intervals. It is evident that the model with CMHA (W/CMHA) achieves significantly better alignment with the ground truth than the model without CMHA (Wo/CMHA), particularly in accurately localizing faults across different hosts and time periods. This indicates that the CMHA mechanism enhances the model's ability to associate task behavior with host status more effectively. 
Table~\ref{tab:full_results} shows the results of the ablation study. We can see that the complete FT-MoE has the best performance compared to Wo/EAGate, Wo/OnlineTuning and Wo/CMHA.

\bibliography{sample}
